\def\year{2022}\relax
\documentclass[letterpaper]{article} 
\usepackage{aaai22}  
\usepackage{times}  
\usepackage{helvet}  
\usepackage{courier}  
\usepackage[hyphens]{url}  
\usepackage{graphicx} 
\usepackage{natbib}  
\usepackage{caption} 
\DeclareCaptionStyle{ruled}{labelfont=normalfont,labelsep=colon,strut=off} 
\usepackage{algorithm}
\usepackage{algorithmic}

\usepackage{newfloat}
\usepackage{listings}
\floatstyle{ruled}
\newfloat{listing}{tb}{lst}{}
\floatname{listing}{Listing}
\nocopyright
\title{Towards Human-Centred Explainability Benchmarks For Text Classification}
\author{
    Viktor Schlegel, Erick Mendez-Guzman and Riza Batista-Navarro
}
\affiliations{

    Department of Computer Science, University of Manchester\\
    Manchester, United Kingdom\\
    \{viktor.schlegel, erick.mendezguzman, riza.batista\}@manchester.ac.uk

}

\begin{document}

\maketitle

\begin{abstract}
Progress on many Natural Language Processing (NLP) tasks, such as text classification, is driven by objective, reproducible and scalable evaluation via publicly available benchmarks. However, these are not always representative of real-world scenarios where text classifiers are employed, such as sentiment analysis or misinformation detection.

In this position paper, we put forward two points that aim to alleviate this problem. First, we propose to extend text classification benchmarks to evaluate the \emph{explainability} of text classifiers. We review challenges associated with objectively evaluating the capabilities to produce valid explanations which leads us to the second main point: We propose to ground these benchmarks in human-centred applications, for example by using social media, gamification or to  learn explainability metrics from human judgements.

\end{abstract}

\section{Introduction}
Recent advancements in deep learning research have yielded impressive results for many 
Natural Language Processing (NLP) tasks,
with text classification~\cite{kowsari2019text,otter2020survey} being no exception.
A popular application domain for text classification systems is social media, as it features
many high-stake use cases~\cite{minaee2021deep}. 
Deep neural networks have been used for suicidal ideation detection and depression identification~\cite{sawhney2018exploring,tadesse2019detection}, analysing public perceptions of physical distancing interventions during the COVID-19 outbreak~\cite{raamkumar2020use}, and identification of modern slavery~\cite{Guzman2022RaFoLa:Labour}. 

Their impressive performance can be attributed to two factors: on the one hand, conceptually simple general-purpose neural architectures that provide expressive contextualised embeddings optimised on large corpora \cite{Devlin2018} are readily available to be fine-tuned to a target application domain. On the other hand, the existence of these standardised approaches facilitates the development of benchmarks \cite{Thorne2018,rajpurkar2016squad,Wang2019,Wang2019a}, that allow us to perform scalable, automated and reproducible evaluation. This, in turn, fosters healthy competition, driven by advantages associated with developing approaches that perform best on these benchmarks, such as increased visibility in the research community  \cite{Linzen2020,Church2019AEvaluation}. In this setting, progress can naturally only be measured (and thus achieved) towards those objectives that are explicitly modelled with corresponding performance metrics as the target of evaluation. This development can be summarised with the proverb ``you reap what you sow'': providing the research community with a competitive and accessible benchmark that evaluates some metric \emph{A} will result in approaches that are explicitly optimised for \emph{A}, gradually pushing \emph{A}'s state-of-the-art measure.

The vast majority of popular benchmarks focus on intrinsic task performance only, reflected in metrics such as accuracy \cite{Ribeiro2020}. 
This may explain more recent findings which suggest that evaluated approaches tend to overfit to these benchmarks by relying on spurious correlations in data in favour of reaching a high score, rather than successfully solving the represented task \cite{gururangan2018annotation}. This is evidenced by their moderate generalisation performance to examples drawn outside of a benchmark's distribution \cite{Schlegel2020a}, or their brittleness towards adversarial inputs \cite{Wallace2019}.
These objectives are of vital importance in real-world scenarios, however. For example, utilising a static dataset to evaluate a tool that detects misinformation on social media contradicts the fact that language use in social media changes over time \cite{eisenstein2014diffusion}, and contemporary events may lead to claims regarding completely new topics. Similarly, an automated tool should be robust to perturbations in the lexical form of how misinformation is conveyed.

Different solutions have been proposed to address these issues, such as ``training set-free benchmarks'' to prevent overfitting to benchmark data \cite{Linzen2020}, or adversarial challenge sets that evaluate the robustness of classifiers to out-of-distribution examples \cite{mccoy2019right}. We propose to take a complementary route and to focus on the capability of classifiers to explain their predictions. In the context of the present position paper, explainability refers to the process of revealing the inner workings behind a model prediction for a single input text \cite{gunning2019xai,danilevsky2020survey}. In contrast, interpretability refers to a characteristic of a model that makes its global logic easily understandable by end-users \cite{atanasova2020diagnostic}. We discuss why explanations are important, what makes explanations ``good'' and how their quality is evaluated and conclude with a set of specific research questions for combining human-grounded evaluation and large-scale objective evaluation in order to make these benchmarks reflective of real-world application scenarios.

\section{Why are Explanations Important?}
The first, perhaps more straightforward use case for explanations is their capability to contextualise predictions. Explanations are necessary for the current generation of state-of-the-art solutions to text classification problems, because
unlike other algorithms, such as those based on symbolic rules, e.g. decision trees, or human-interpretable features, it is not directly possible to predict the behaviour of a neural network-based classifier (the \emph{what}) or to explain it (the \emph{why}) by looking at its performed computations (the \emph{how}). Neural networks are agnostic to input data (e.g. the same algorithm can be used to classify text or detect cats in images); the properties of the resulting models are largely determined by the data they are optimised on. This is not necessarily a problem: if there is access to data that is fully representative of a well-defined task, as is the case with synthetic datasets \cite{Weston2015}, and a model that solves this task perfectly, one might argue that understanding its behaviour is secondary given the reliability of its performance. 

As previously argued, this assumption appears to not hold for many text classification tasks due to the dynamic and compositional nature of language, and can be seen by the lack of out-of-distribution generalisation. In this scenario, an explanation can provide end-users the necessary context to trust or challenge the prediction \cite{Glass2008TowardAgents}. For example, a nonsensical explanation to a social media post classified as misinformation could hint at the fact that the prediction is wrong. In this context, explanations serve the same purpose as many other tools aimed at providing additional context, such as with model cards \cite{Mitchell2019ModelReporting}, behavioural testing suites \cite{Ribeiro2020} or interpretability methods \cite{Belinkov2019a}. Different from those, explanations are arguably the easiest to understand for end users as they do not require any technical knowledge.

Beyond their use as a diagnostic tool, explanations can help to evaluate the classifier itself. This is intuitive: being able to explain something is a sufficient but not necessary indicator to understand it---to explain something you need to understand it first and good performance on the harder task of producing a good explanation implies robust performance on the classification task. This intuition is supported by the  fact that jointly learning to both predict and explain, results in more robust classifiers that are less susceptible to weaknesses such as reliance on spurious correlations \cite{Rajagopal2020,Dua2020BenefitsComprehension,atanasova2020diagnostic,lei2016rationalizing}.

Therefore, objective and large-scale evaluation of classifiers' capabilities to produce explanations is desirable, and benchmarks that require approaches to explicitly optimise these capabilities, can contribute to the development of robust classifiers suitable for real-world applications. In the following sections we discuss why such benchmarks have yet to gain traction.

\section{What Makes a Good Explanation?}
Even in practical terms, what constitutes a valid explanation is an open question. On a high level, different desiderata for evaluating explanations have been proposed: \emph{plausibility}, or human-likeness describes whether humans are likely to be convinced by a machine-generated explanation. Put simply, an explanation is plausible if a majority of humans exposed to the explanation agree that it explains the prediction well.
Furthermore, \emph{faithfulness}, also referred to as fidelity, describes the extent to which explanations inform model predictions~\cite{ross2017right}, which is important, because a classifier's plausible explanation should in fact be reflected in its prediction.

Different approaches have been proposed for explaining the outcome of neural text classifiers. Explanations by example exemplify a classification by presenting most similar examples in the training data \cite{Han2020}. Counter-factual explanations, rooted in the causal reasoning framework \cite{hendricks2018generating,garg2019counterfactual} identify which parts of the input need to be changed in order to arrive at a different explanation. Scientific explanations aim to link the input to the predicted label via reasoning chains and background knowledge \cite{Jansen2018WorldTree:Anthology}, Finally, \emph{rationales} are text snippets that justify or warrant the predicted class.

In its simplest form, these are based on building connections between the input text and output labels and quantifying how much each element (words or \emph{n}-grams) contributes to the final prediction~\cite{atanasova2020diagnostic}. In the context of deep learning-based methods for text classification, these extractive post-hoc explanation methods commonly take the form of token-level importance scores~\cite{danilevsky2020survey}. While \textit{soft} scoring models assign soft weights to token representations, and so one can extract highly weighted tokens as explanations, \textit{hard} selection mechanisms discretely extract snippets from the input text as explanations~\cite{atanasova2020diagnostic,danilevsky2020survey}. Relaxing the requirement that rationales must be part of the input, free-form rationales take the form of short texts aimed at justifying the label \cite{liu2018towards}. These can be generated with Natural Language Generation approaches, such as Sequence-to-Sequence models \cite{costa2018automatic,rajani2019explain}.

















\section{How is Explainability Evaluated?}
Concerning evaluation, there is little consensus on how to assess the quality of these explanations either for deployment or benchmarking~\cite{doshi2017towards}. Current evaluation approaches typically fall into two categories. The first uses some formal definition of explainability desiderata as a quantifiable proxy for explanation quality and requires no human experiments~\cite{nguyen2018comparing,carton2020evaluating}. The second evaluates explainability in the context of an application using human experiments either for the practical application or a simplified version of it~\cite{lertvittayakumjorn2019human,hase2020evaluating}.

\paragraph{Proxy-based Evaluation}
Model-generated explanations have typically been evaluated by automatic proxy measures of similarity with human explanations~\cite{hase2020evaluating}. In this context, the term ``proxy'' denotes that an objective metric is used to---perhaps imperfectly---represent a more abstract concept, such as plausibility. For the discrete case, researchers have measured agreement using phrase-matching metrics such as BLEU and MAXSIM~\cite{camburu2018snli,clinciu2021study}, Intersection-over-Union (IOU) on a token level to derive precision, recall and F1-scores~\cite{deyoung2019eraser}. Meanwhile, for soft token scoring models, some studies have used the Area Under the Precision-Recall curve (AUPRC) constructed by sweeping a threshold over token scores~\cite{samek2016evaluating}.

While these proxy metrics are suitable for automated and scalable evaluation, there is no guarantee that they represent human intuition well. At their core, these metrics are evaluating lexical overlap and might give unreasonably high scores when predicted rationales are lexically similar yet semantically different or unreasonably low scores in the converse case. For text generation tasks, this results in poor correlation between human ratings and scores under these metrics \cite{Wiseman2017ChallengesGeneration,Ma2019ResultsChallenges}. Additionally, for generated explanations, trivial baselines, such as input copying, may outperform bespoke approaches \cite{Krishna2021HurdlesAnswering}. 

Furthermore, even though the evaluation metrics cited above provide a good starting point, they do not reveal model behaviour as they evaluate plausibility but not faithfulness~\cite{hase2020leakage,carton2020evaluating}, because they do not reveal whether the produced explanations actually informs the classifier's prediction. Despite the fact that the best way to measure explanation faithfulness for text classification remains an open question, most researchers investigating it have defined that a faithful explanation should have high sufficiency and comprehensiveness~\cite{deyoung2019eraser}. Sufficiency measures how well explanations can provide nearly the same prediction as the whole input text, while comprehensiveness assesses how well machine-generated explanations include all relevant tokens~\cite{liu2018towards}.

More recently, literature presenting contradictory findings of using proxy-based metrics for evaluating machine-generated explanations, has emerged \cite{carton2020evaluating,lertvittayakumjorn2019human}. Previous research has established that human explanations do not necessarily have high sufficiency and comprehensiveness, suggesting they may not be treated as ground truth baselines~\cite{carton2020evaluating}.

The issue has grown in importance in light of recent evidence that indicates that automated metrics are highly model-dependent, which makes it challenging to compare explanations from different text classification models~\cite{clinciu2021study}. Moreover, metrics for assessing explanation faithfulness are unstable compared to model performance. For instance, sufficiency and comprehensiveness may continue to fluctuate significantly even when model accuracy has stabilised~\cite{carton2020evaluating}.


\paragraph{Human-grounded Evaluation}
With a concrete application in mind, the best way to assess model explanations is to evaluate them for the specific task. If such evaluation with end-users is not feasible, simpler human-subject experiments may be a good alternative if they maintain the essence of the target application.

To date, several studies have begun to examine the use of human-grounded evaluation methods to assess explanation quality for NLP applications~\cite{mohseni2018human,nguyen2018comparing,lertvittayakumjorn2019human,mohseni2021quantitative}. Together, these studies indicate that human experiments can help to evaluate whether explanations can expose irrational behaviour of a text classifier, e.g. the exploitation of dataset-specific spurious correlations~\cite{lertvittayakumjorn2019human}, assess whether explanations are faithful concerning the predicted class~\cite{mohseni2018human,mohseni2021quantitative} or to investigate uncertain predictions~\cite{lertvittayakumjorn2019human}. Human-centred experiments can also help to measure the users' understanding of explanations, establish the usefulness of explanations for a specific task \cite{Paleja2021TheTeaming}, and assess user-reported trust as a proxy for explanation goodness \cite{lertvittayakumjorn2019human,lage2019human,kocielnik2019will}.


\section{How to Benchmark Explainability}
We have reviewed the challenges of evaluating text classification systems robustly, objectively and at scale. Objective evaluation is exacerbated by spurious correlations in static benchmarks and changing requirements between training data and real-world application scenarios. To detect and alleviate these issues, we propose to expand text classification benchmarks by a harder task, \textbf{explanation reconstruction}. This forms a sufficient requirement for successful text classification, while doing so ``for the right reasons'', rather than relying on superficial heuristics and shortcuts. Specifically, we discuss the tasks of rationale extraction and generation as conceptually simple yet flexible candidates. In this section, we outline the necessary steps to transition from performance-focussed benchmark suites to those that explicitly evaluate explainability. 

In the \textbf{short-term}, we argue that rationales need to be incorporated into large-scale classification benchmarks. Conceptually, both generating and extracting rationales are well-explored supervised learning tasks (natural language generation and span extraction, respectively), so any classification benchmarks and classifiers they evaluate can be feasibly enriched by these tasks. 
\emph{For new benchmarks}, required training and evaluation data can be collected by crowd-workers during annotation for the main task due to the conceptual simplicity of rationales, as they do not require additional knowledge, such as a training set to draw analogies from or additional resources to form reasoning chains. In other words, providing a justification does not significantly increase the cognitive load of annotators, because they already perform this task when choosing the correct label, albeit implicitly \cite{Zaidan2007UsingAnthology}.
\emph{For existing benchmarks}, rationale annotations can be gathered post-publication \cite{Dua2019a}.

Evaluation of explainability in such benchmarks can be carried out with the existing proxy-based metrics based on overlap with human-annotated rationales.

In the \textbf{longer term}, however, proxy-based metrics only reflect general desiderata of explainability \cite{lertvittayakumjorn2019human}, and are not aimed at establishing the \emph{usefulness} of produced explanations. By usefulness we mean the level of utility an explanation brings in addition to the raw prediction, when supporting decision-making. For example, a human might report an explanation to a misleading claim to be plausible and faithful. However, this is not necessarily reflective of whether the same human would refrain from sharing the claim on social media. 
Thus, to better reflect real-world human-centred applications in benchmarks, requirements for valid explanations for different tasks and for different user groups need to be gathered. As argued previously, human-centred experiments are best to establish and repesent these requirements.
Unfortunately, these evaluation methods do not exhibit any of the features that make benchmarks popular: Due to the human involvement they are not automated, and they might not be easily scalable and reproducible due to the effort required. Every newly developed approach will require a costly re-run of the human-centred evaluation.



With the need of objective explanation evaluation and in light of  the challenges associated with existing proxy-based approaches, we propose to develop evaluation metrics based on task-specific signals provided by humans, effectively simulating expensive human-centred experiments. We propose three possible ways to achieve and highlight associated practical questions that need to be considered.

\emph{Gamification:} Gamification (``Beat the AI'') or dynamic benchmarks have been  proposed more recently to combat the reliance on dataset-specific artefacts in static benchmarks \cite{Nie2019,Kiela2021Dynabench:NLP,Wallace2019TrickAnswering}. In these settings, examples would only be accepted if they can ``fool'' some pre-trained classifier to produce the wrong label, while the majority of humans would still annotate the correct one. The game-like nature of the annotation task provides additional motivation, at least in the short term \cite{Hamari2014DoesGamification}.
Similarly, this annotation paradigm can be extended to continuously evaluate explanations generated by optimised models, for example by accepting only those challenging examples, where an explanation generated by a system would be voted as low-quality by majority of human annotators, effectively serving as a human-in-the-loop explanation benchmark. 

\emph{Learned explainability evaluation metrics:} More recently, overlap-based metrics have been complemented by embedding-based metrics in an effort to improve the correlation between automatically measured and human-judged quality of generated text \cite{Zhang2019BERTScore:BERT,Sellam2020BLEURT:Generation}. Similar metrics can be developed to directly learn plausibility, faithfulness or usefulness of explanations based on human judgements, effectively approximately replicating human-centred experiments. Predicting these scores directly given an explanation, has the further advantage that explanations can potentially be evaluated \emph{without gold standard reference explanations}. However, it is worth investigating, whether the learned explainability metrics generalise beyond the training/evaluation data or whether the metrics themselves are susceptible to spurious correlations. Finally, overlap-based metrics have the advantage of being interpretable. It is unclear whether this property is desired from learned metrics.

\emph{Use of Social Media:} Finally, the dynamics of social media can be helpful to approximate how \emph{useful} or convincing generated explanations are. \citet{Zellers2020EvaluatingUse} post machine-generated explanations to the \textsc{RedditAdvice} forum and approximate their usefulness by the number of ``upvotes'' received. Similarly, researchers can expose generated explanations to the public by making use other fora that have voting, liking or sharing mechanisms. For example, the utility of fact verification tools can be approximated by automatically identifying misleading tweets and commenting with generated explanations why they are misleading. Different explanation methods can then be compared by the number of retweets they can accumulate or by how much they manage to dampen the expected spread of the original claim. Studies like these will need to consider the trade-off between impartiality and potential harm. On the one hand prefacing generated explanations with the fact that they are machine-generated can have an impact on evaluation, while on the other hand, ``incorrect explanations'' could cause harm.


We hope this position paper will contribute to the ongoing dialogue around incorporating human-centred, application-grounded scalable evaluation of explainability capabilities into predominantly performance-based benchmarking of text classification approaches. This will contribute to the development of robust benchmarks that are not easily exploitable firstly, and secondly, are more representative of humans' needs in real-world scenarios.

\section{Acknowledgements}
We thank the reviewers for their feedback. This work was partially funded
by the European Union’s Horizon 2020 research and innovation action programme, via the AI4Media Open Call \#1 issued and executed under the AI4Media project (Grant Agreement no. 951911).


\bibliography{references_emg,manually}
\end{document}